\documentclass[runningheads]{llncs}

\usepackage{eccv}

\usepackage{eccvabbrv}

\usepackage{graphicx}
\usepackage{booktabs}

\usepackage[accsupp]{axessibility}  % Improves PDF readability for those with disabilities.

\usepackage{hyperref}

\usepackage{orcidlink}

\usepackage{url}            % simple URL typesetting
\usepackage{booktabs}       % professional-quality tables
\usepackage{amsfonts}       % blackboard math symbols
\usepackage{nicefrac}       % compact symbols for 1/2, etc.
\usepackage{microtype}      % microtypography
\usepackage{xcolor}         % colors
\usepackage[linesnumbered,ruled,vlined]{algorithm2e}
\usepackage{amssymb}
\usepackage{amsmath}
\usepackage{pifont} % http://ctan.org/pkg/pifont
\usepackage{algpseudocode}
\usepackage{wrapfig}
\usepackage{multirow}
\usepackage{color, colortbl}
\usepackage{cuted}
\definecolor{Gray}{gray}{0.9}

\begin{document}

% \title{Supplementary Materials for Context Diffusion: In-Context Aware Image Generation }
\title{Context Diffusion: In-Context Aware Image Generation}
\titlerunning{Context Diffusion: In-Context Aware Image Generation}


\author{
% \hspace{4mm}
Ivona Najdenkoska\textsuperscript{1,2}\thanks{Work done during an internship at Meta GenAI. Correspondence at \email{i.najdenkoska@uva.nl}.}
\hspace{3mm} Animesh Sinha\textsuperscript{1}
\hspace{3mm} Abhimanyu Dubey\textsuperscript{1}\\
\hspace{0mm} Dhruv Mahajan\textsuperscript{1}
\hspace{0mm} Vignesh Ramanathan\textsuperscript{1}
\hspace{0mm} {Filip Radenovic}\textsuperscript{1}
}

\institute{\textsuperscript{1}Meta GenAI \hspace{5mm} \textsuperscript{2}University of Amsterdam}

\authorrunning{I.~Najdenkoska et al.}

\maketitle

\begin{abstract}
\vspace{-.3cm}
We propose Context Diffusion, a diffusion-based framework that enables image generation models to learn from visual examples presented in context. Recent work tackles such in-context learning for image generation, where a query image is provided alongside context examples and text prompts. However, the quality and context fidelity of the generated images deteriorate when the prompt is not present, demonstrating that these models cannot truly learn from the visual context. To address this, we propose a novel framework that separates the encoding of the visual context and the preservation of the desired image layout. This results in the ability to learn from the visual context and prompts, but also from either of them. Furthermore, we enable our model to handle few-shot settings, to effectively address diverse in-context learning scenarios.
Our experiments and human evaluation demonstrate that Context Diffusion excels in both in-domain and out-of-domain tasks, resulting in an overall enhancement in image quality and context fidelity compared to counterpart models.
\end{abstract}
   
\section{Introduction}
\label{sec:intro}
\begin{figure}[t]
  \centering
   \includegraphics[width=\textwidth]{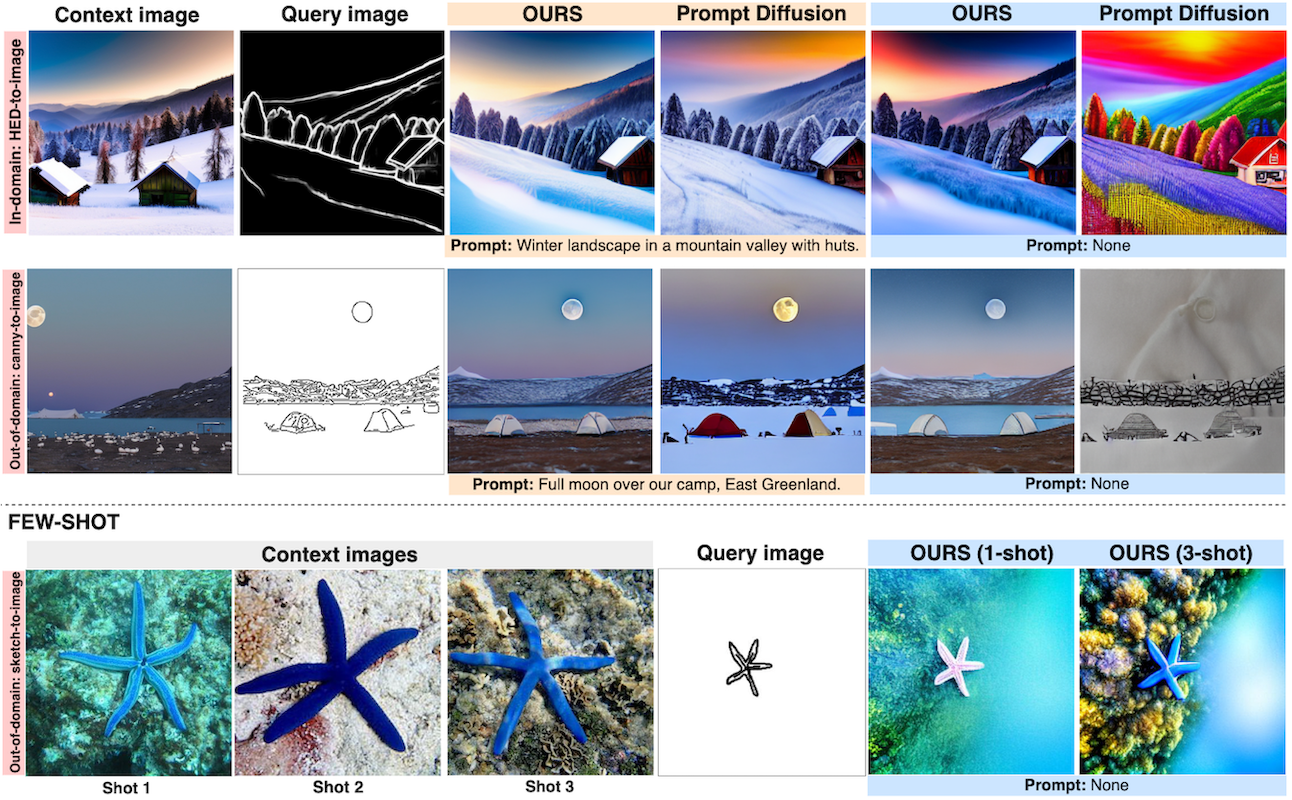}
   \vspace{-0.5cm}
   \caption{
        \textbf{Illustrating in-context aware image generation with Context Diffusion. } \textbf{Top row}: HED-to-image as an in-domain task; \textbf{Middle row}: canny-to-image as an out-of-domain task. Our model enables learning from context with and without prompts. The counterpart model, Prompt Diffusion \cite{wang2023context} cannot leverage the context if the prompt is not provided, hinting at its over-reliance on textual guidance;
    \textbf{Bottom row}: Few-shot setting for sketch-to-image task. More context examples help in learning stronger visual cues, even without prompts.
    }
   \label{fig:examples_teaser}
   \vspace{-0.5cm}
\end{figure}

Generative models are witnessing major advances, both in natural language \cite{brown2020language,chowdhery2022palm,zhang2022opt,wei2021finetuned,ouyang2022training,touvron2023llama} and media generation~\cite{rombach2022high,saharia2022photorealistic,huang2023noise2music,poole2022dreamfusion,brooks2023instructpix2pix,ruiz2023dreambooth,dai2023emu,team2023gemini}.
Large language models in particular have shown impressive in-context learning capabilities \cite{brown2020language,wei2022emergent,tsimpoukelli2021multimodal,alayrac2022flamingo}. This represents the ability of a model to learn from a few samples on the fly, without any gradient-based updates, and extend it to new tasks and domains. However, for generative models in computer vision, learning from context examples remains under-explored. 

The closest line of work that explicitly supports a single image pair as a context example for image generation is Prompt Diffusion \cite{wang2023context}.
% is perhaps the closest line of work that explicitly supports a single source-target image pair as a context example for image generation. 
It builds on the popular ControlNet \cite{zhang2023adding} model which introduced the idea of controllable diffusion models. Specifically, Prompt Diffusion attempts to learn the visual mapping from a source image to a target context image and applies it to a new query image, by also leveraging a prompt for text-based guidance. However, we empirically observed that this model struggles to leverage the context example when the text prompt is absent.
This results in low fidelity to the visual context examples, particularly when the examples are from a different domain than what is seen during training. For instance, if the source-target pair shows specific styles, they cannot be leveraged during inference just from the context examples. 
This is seen in the first row of Figure~\ref{fig:examples_teaser} where Prompt Diffusion is unable to learn the ``snowy'' style from the context unless prompted through text. 
Additionally, it does not trivially support multiple images as context examples, which limits the visual information that can be provided to the model. 

We address these challenges with our proposed \textit{Context Diffusion} model that can (i) effectively learn from visual context examples as well as follow text prompts and (ii) support a variable number of context examples since visual characteristics can be defined with more than a single example. Unlike Prompt Diffusion, our model does not require paired context examples, but just one or more ``target'' context images serving as examples of the desired output and a single query image providing visual structure.
The reason for using solely target examples as visual context is that the source images are derived from the target itself and do not provide any additional information for the task.
Typically, the query image provides guidance for the output layout through edges, depth, segmentation maps, etc. On the other hand, the context examples provide hints for finer details such as styles, textures, color palettes, and object appearances desired in the output image.

Furthermore, it is important to note the difficulty in controlling both aspects of the output image solely through the control mechanisms \cite{zhang2023adding,wang2023context}. The ``control'' part of the model is very effective in capturing high-level structure and layouts. However, fine-grained details are better captured through the conditioning mechanism, as seen in textual inversion \cite{gal2022image}, grounded text-to-image generation \cite{li2023gligen}, and retrieval-augmented image generation \cite{chen2022re, rdm2022blatt}.
To that end, we inject the visual information from the context into the network as text conditioning. 
Different from prior works, we aggregate the sequence of all visual embeddings extracted from the context images and place them alongside the text embeddings in the cross-attention layers of the diffusion model. The ability to aggregate a variable number of context images enables the model to handle few-shot scenarios.
Moreover, these modified cross-attention layers allow a stronger reliance on the visual context which is especially apparent without textual guidance. The layout of the query image is preserved by passing it as a control signal to the network in a similar manner as ControlNet \cite{zhang2023adding}.
  
At inference time, we use a query image to define the target layout and one or more context images to provide finer visual signals, alongside an optional text prompt. 
Our experiments study the generation ability of Context Diffusion for in-domain tasks, such as using HED, segmentation, and depth maps to generate real images and vice versa. We show the flexibility of our model to preserve the structure and layout from the query image and transfer visual signals from the context even when the text prompt is not provided. Moreover, to properly study in-context learning abilities, we experiment with unseen, out-of-domain tasks, such as handling sketches, image editing, and more. 
Furthermore, for such tasks, using multiple images in a few-shot manner yields stronger fidelity of the generated images to the context. 

\paragraph{Contributions.}
(i) We propose Context Diffusion, an in-context aware image generation framework. It enables pre-trained diffusion models to leverage visual context examples to control the fine-grained details of the output image, alongside a query image that defines the layout and an optional text prompt. (ii) We enable the use of multiple context images as ``few-shot'' examples for image generation. To the best of our knowledge, this is the first work to explore such a ``few-shot'' setup for in-context-aware image generation. (iii) We conduct extensive offline and online (human) evaluations that show that our framework can handle several in-domain and out-of-domain tasks and demonstrates improved performance over the prior work.

\section{Related Work}
\label{sec:related_work}

\paragraph{Diffusion-based Image Generation.}
Recent advancements in diffusion models, first introduced in \cite{pmlr-v37-sohl-dickstein15} have exhibited huge success in text-to-image generation tasks \cite{ramesh2021zero,ho2020denoising,ramesh2022hierarchical,saharia2022photorealistic,dai2023emu}. Enhancements have been achieved through various
training \cite{rombach2022high,saharia2022photorealistic,dai2023emu} and sampling \cite{song2021denoising, lu2022dpm,wang2023patch} techniques. For instance, DALLE-2 \cite{ramesh2022hierarchical} proposed an architecture encompassing several stages, by encoding text with CLIP \cite{clip_icml21} language encoder and decoding images from the encoded text embeddings, followed by Imagen \cite{saharia2022photorealistic} which showed that up-scaling the text encoder largely improves the text fidelity. 
Furthermore, the Latent Diffusion Model (LDM) \cite{rombach2022high} investigated the diffusion process by applying it to a low-resolution latent space and even further improved the training efficiency. 
However, all these models only take a text prompt as input, which restricts the flexibility of the generation process as it requires extensive prompt engineering to obtain the desired image outputs.

\paragraph{Controllable Image Generation.}
Besides the text prompts, adding more control to the image generation process helps overall customization and task-specific image generation. Recent text-conditioned models focus on adjusting models by task-specific fine-tuning \cite{ruiz2023dreambooth,gal2022image,chen2023suti, wei2023elite, gal2023encoder}, injecting conditioning maps, like segmentation maps, sketches or key-points \cite{li2021imgsyn,avrahami2023spatext, bashkirova2023masksketch, gafni2022make, zhang2023adding, qin2023unicontrol, fan2023frido, zhao2024uni}, or exploring editing abilities \cite{meng2022sdedit,hertz2022prompt,mokady2023null,brooks2023instructpix2pix,goel2023pair}. For instance, SpaText \cite{avrahami2023spatext} is using segmentation maps where each region of interest is annotated with text, to better control the layout of the generated image. Models like GLIGEN \cite{li2023gligen} inject grounding information, such as bounding boxes or edge maps, into new trainable layers via a gated cross-attention mechanism. ControlNet \cite{zhang2023adding}, as a recent state-of-the art in controllable image generation presents a general framework for adding spatial conditioning controls. 
UniControl \cite{qin2023unicontrol} extends ControlNet by unifying various image map conditions into a single image generation framework. 
Other works, such as Re-Imagen \cite{chen2022re} and RDM \cite{rdm2022blatt}, employ retrieval for choosing images given a text prompt, for controlling the generation process. 

Our approach differs from these models in several aspects. We support learning from in-context visual examples as an addition to the textual prompts and query images. This allows learning new tasks using the visual context only, which yields a more flexible framework. 
Additionally, we use only a few of the image maps considered by ControlNet and UniControl for training, namely HED, segmentation, and depth maps, and demonstrate the generalization ability to the other visual controls \ie query images.

\paragraph{In-Context Learning in Image Generation.} Although in-context learning is vastly explored both in language-only \cite{brown2020language,wei2021finetuned,kojima2022large,wei2022emergent} and visual-language models \cite{alayrac2022flamingo,tsimpoukelli2021multimodal,li2023blip,koh2023grounding}, its application is lagging behind in image generation. 
Bar \etal \cite{bar2022visual} investigate visual prompting for image generation, followed by Painter \cite{wang2023images} which incorporates more tasks to construct a generalist model.  
Regarding in-context abilities in diffusion models, Prompt Diffusion \cite{wang2023context} presents such a framework that extends the control capability of ControlNet \cite{zhang2023adding} for in-context image generation. 
They consider a vision-language prompt encompassing a source-target image pair and a text prompt, which is used to jointly train the model on six different tasks.
However, Prompt Diffusion only shows good performance when both the context images and prompt are present. In case the text prompt is not present, the model exhibits deteriorating performance, suggesting its inability to learn efficiently from the visual examples, as shown in Figures \ref{fig:examples_teaser}, \ref{fig:examples_in_domain}, \ref{fig:examples_edits}, \ref{fig:out_domain_examples}, \ref{fig:examples_no_prompts}, and \ref{fig:examples_no_context}.
Different from them, we aim to develop a model able to generate images of good quality even when only one of the conditions (visual context or text prompt) is present. 

Another work tackling image generation with visual examples is Prompt-Free Diffusion \cite{xu2023prompt}. It focuses only on having an image as a context \ie a visual condition, while completely removing the ability to process textual prompts. This is the major difference compared to our Context Diffusion, since we aim to support both scenarios: having the context images and/or text prompts. 
Additionally, none of these related works consider settings with multiple examples in context, namely, few-shot scenarios. We propose a framework that can handle a variable number of context images, helpful for enriching the visual context representation.

\section{Methodology}
\label{sec:method}

\subsection{Preliminaries}
Diffusion models are a class of generative models that convert Gaussian noise into samples from a learned data distribution via an iterative denoising process. In the case of diffusion models for text-to-image generation, starting from noise $z_t$, the model produces less noisy samples $z_{t-1}, \dots, z_0$, conditioned on caption representation $\mathbf{c}$ at every time step $t$. 

To learn such a model $f_\theta$ parameterized by $\theta$, for each step $t$, the diffusion training objective $\mathcal{L}$ solves the denoising problem on noisy
representations $z_t$, defined as follows:
\begin{equation}
\label{eq:objective}
    \min_\theta \mathcal{L} = \mathrm{E}_{z, \epsilon \backsim \mathcal{N}(0, 1),t } \bigr[\parallel \epsilon - f_\theta(z_t, t, \mathbf{c})  \parallel^2_2 \bigr],
\end{equation}
% where $f_\theta$ is the denoising autoencoder.
With large-scale training, the model $f_\theta$ is trained to denoise $z_t$ based on text information as the main source of control. 

To enable more control over the generation process, we follow the ControlNet setup \cite{zhang2023adding}, for encoding the layout of the desired output via a query image as visual control. Note that we use \textit{query image} interchangeably with \textit{visual control}.
In this paper, we extend the $\mathbf{c}$ representation in Eq. (\ref{eq:objective}), by adding image examples as additional guidance besides the text prompt.
Namely, we inject visual embeddings obtained by a pre-trained vision encoder $f_{\text{img}}$ with fixed parameters, in a similar fashion as the text embeddings.

\subsection{Context Diffusion Architecture}
The model $f_\theta$ is essentially a UNet architecture \cite{ronneberger2015u}, with an encoder, a middle block, and a skip-connected decoder. 
\begin{figure}[t]
  \centering
   \includegraphics[width=\textwidth]{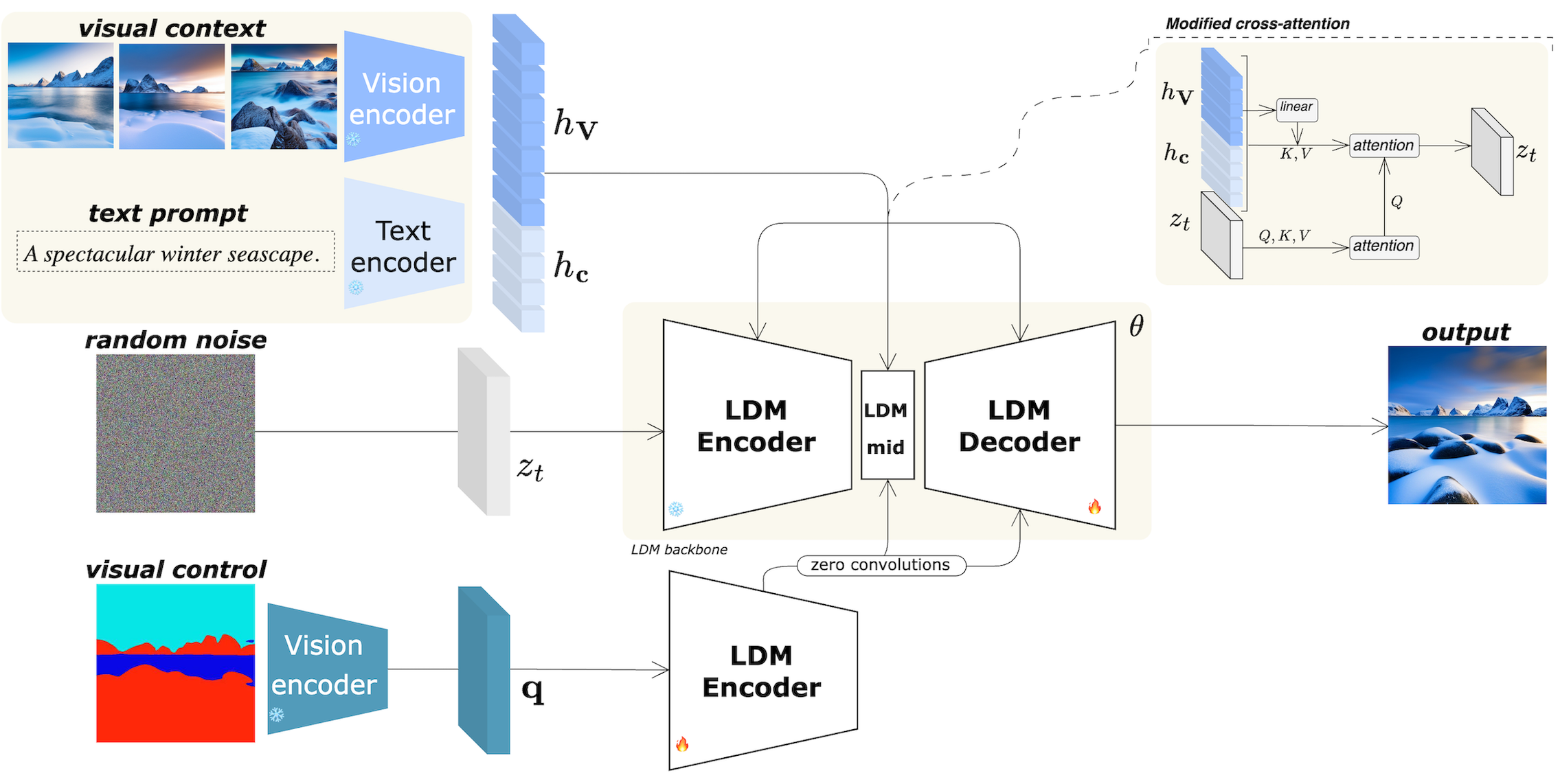}
   \vspace{-0.2cm}
   \caption{\textbf{Architecture of Context Diffusion.} It consists of several modules: vision and text encoders for encoding the text prompt and visual context respectively, an LDM backbone for handling the image generation process, and an additional LDM encoder for processing the query image as a visual control. 
   Note that here we show three visual context examples, however, the model is trained using a variable number of such examples. }
   \label{fig:architecture}
   \vspace{-0.5cm}
\end{figure}
These modules denoted as LDM encoder, mid, and decoder in Figure \ref{fig:architecture}, are built out of standard ResNet \cite{he2016deep} and Transformer blocks \cite{vaswani2017attention} which contain several cross-attention and self-attention mechanisms. The core of conditional-diffusion models is encoding the conditional information \cite{rombach2022high}, based on which $z_t$ is generated at a given time step $t$. We differentiate two types of such conditional information: the visual context $\mathbf{V}$ encompassing images and the text prompt $\mathbf{c}$, to define our conditioning information: $\mathbf{y} = (\mathbf{c}, \mathbf{V})$,
where $\mathbf{V} = [\mathbf{v}_1, \dots \mathbf{v}_k]$ and $k$ denotes the number of images. Additionally, we consider a visual control image, \ie, the query image that serves to define the layout of the output denoted as $\mathbf{q}$. 

\paragraph{Prompt encoding.} To perform the encoding of the textual prompt $\mathbf{c}$ we use a pre-trained language encoder $f_{\text{text}}$ with fixed parameters to obtain the embeddings. Particularly, we obtain $\mathbf{h}^\mathbf{c} = \{h^\mathbf{c}_0,\dots,h^\mathbf{c}_{N^\mathbf{c}}\} = f_{\text{text}}(\mathbf{c})$, where $N^\mathbf{c}$ is the number of text tokens, $h^\mathbf{c}_i \in \mathbb{R}^{d^\mathbf{c}}$ and $d^\mathbf{c}$ is the dimensionality of the textual token embeddings.
\paragraph{Visual context encoding.}  
We hypothesize that the visual context $\mathbf{V}$ should be at the same level of conditioning as the textual one. Therefore, we follow a similar strategy for encoding the visual context, by using a pre-trained, fixed image encoder $f_{\text{img}}$. 
Given a visual context $\mathbf{V}$ consisting of $k$-images, we encode each image $\mathbf{v}_i$ as $\mathbf{h}^{\mathbf{v}_i} = \{h^{\mathbf{v}_i}_0,\dots,h^{\mathbf{v}_i}_{N^{\mathbf{v}}}\} = f_{\text{img}}({\mathbf{v}_i})$, where $N^\mathbf{v}$ is the number of tokens per image, $h^{\mathbf{v}_i} \in \mathbb{R}^{d^\mathbf{v}}$ and $d^\mathbf{v}$ is the dimensionality of the visual token embeddings. 
The final representation of the visual context is obtained by simply summing the corresponding visual tokens of all $k$-images, where $k \in \{1, 2, 3\}$, yielding $ \mathbf{h}^\mathbf{V} = \sum_{i=1}^k \mathbf{h}^{\mathbf{v}_i}$.
Then, we add a linear projection layer to map the visual embedding dimension $d^\mathbf{v}$ to the language dimension $d^\mathbf{c}$.
\paragraph{Modified cross-attention.} Given the standard cross-attention block in LDMs, defined with queries $Q$, keys $K$, and values $V$, the noisy representation $z_t$ is used as a query, whereas the text encoding $\mathbf{h}^\mathbf{c}$ is used as a representation of the keys and values, as follows: 
\begin{equation}
    z_t = z_t + \text{CrossAtt}(Q=z_t, K=V=\mathbf{h}^\mathbf{c}).
    \label{eq:cross_att}
\end{equation}
Our framework is slightly different from this definition since we also consider visual information in the conditioning. Therefore, after obtaining both visual and textual embeddings we concatenate them to obtain
$[\mathbf{h}^\mathbf{c}, \mathbf{h}^\mathbf{V}]$, illustrated in the bottom left corner of Figure \ref{fig:architecture}. 
Thus the input to the cross-attention block in (\ref{eq:cross_att}) changes as follows:
\begin{equation}
    z_t = z_t + \text{CrossAtt}(Q=z_t, K=V=[\mathbf{h}^\mathbf{c}, \mathbf{h}^\mathbf{V}]).
\end{equation}

\paragraph{Visual control encoding.} 
To enable the ingestion of the query image as visual control, we follow ControlNet setup \cite{zhang2023adding}. First, the image is encoded using a few convolutional layers. Then, a copy of the LDM encoder is used to process the encoded query image $\mathbf{q}$. This trainable LDM encoder copy is connected to the original LDM backbone using zero convolution layers, as shown in Figure \ref{fig:architecture}.

\subsection{Multi-task Training Procedure}
We use a pre-trained image generation model to adapt it with visual context injection. We use the original denoising objective defined in (\ref{eq:objective}), with $\mathbf{q}$ being the query image and the modified conditioning information $\mathbf{y}$: 
\begin{equation}
\label{eq:objective_1}
    \min_\theta \mathcal{L} = \mathrm{E}_{z, \epsilon \backsim \mathcal{N}(0, 1),t } \bigr[\parallel \epsilon - f_\theta(z_t, t, \mathbf{y}, \mathbf{q})  \parallel^2_2\bigr].
\end{equation}
To train with this objective, we use a collection of tasks for joint end-to-end training, similar to \cite{wang2023context}. Different from them, we use a visual context sequence consisting of a $k$-images and an optional text prompt, together with a query image. Specifically, $k$ is randomly chosen at batch construction. The goal of such training is to leverage any visual characteristics from a variable number of context images and to apply them along with the text prompt to the query image. 

\section{Experiments}
\label{sec:experiments}

\subsection{Experimental Setup}
\paragraph{Datasets.} To train our model, we use a dataset that consists of 310k synthetic images and caption pairs, similar to Prompt Diffusion \cite{wang2023context}. 
Following their training setup and code implementation, we extract three image maps: HED, segmentation, and depth maps from the training images. 
During training, for map-to-image tasks, the image maps serve as queries, and real images are used for visual context, while for image-to-map tasks, the real images serve as queries, and image maps are used for visual context. Note that the prompts and visual context are usually related and describe a similar conditioning signal.

At inference time, we use the test partition of the dataset to test the ability of the model to learn from context. To demonstrate the generalization abilities of Context Diffusion to out-of-domain tasks, we extract other image maps, such as normal maps, canny edges, and scribble maps. Also, we consider editing tasks by using real images as queries. To further test the generalization abilities, we utilize hand-drawn sketches from the Sketchy dataset \cite{sketchy2016sangkloy}, where the sketch is the query image, and the real images are the visual context. This dataset does not provide captions, therefore we construct text prompts using a template: ``A professional, detailed, high-quality image of \textit{object name}'', following \cite{zhang2023adding}.

\begin{figure}[t]
  \centering
   \includegraphics[width=\textwidth]{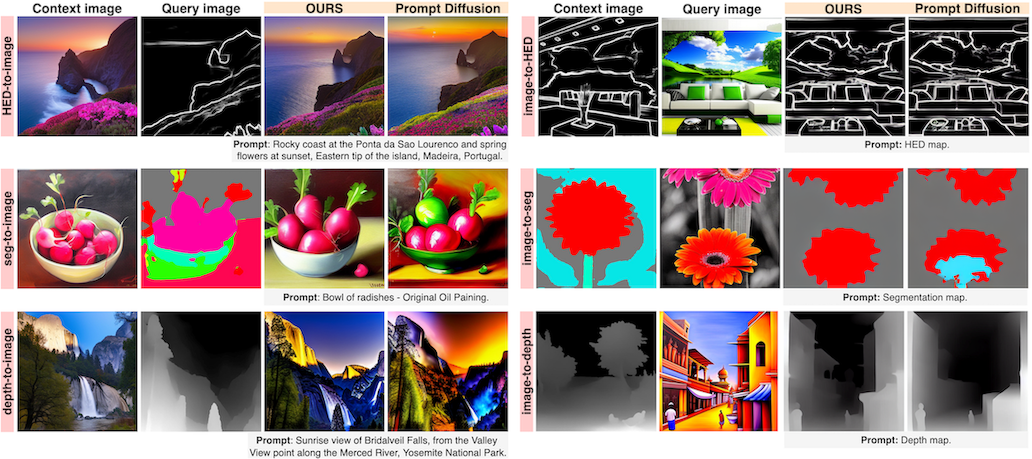}
   \vspace{-0.5cm}
   \caption{
        \textbf{In-domain comparison to Prompt Diffusion~\cite{wang2023context}:} 
        Examples of \{HED, segmentation, depth\}-to-image as forward tasks and image-to-\{HED, segmentation, depth\} as reverse tasks, with both visual context and prompt (C+P) given as conditioning information.
    }
    \vspace{-0.5cm}
   \label{fig:examples_in_domain}
\end{figure}

\paragraph{Implementation Details.} The backbone of our model follows a vanilla ControlNet architecture, initialized in the same way as~\cite{wang2023context}.
We train the model using the data setup explained above. In particular, only the encoder of the LDM backbone is kept frozen and its copy which processes the query image is trained. For the encoding of the context images and prompts, we use frozen CLIP ViT-L/14 \cite{clip_icml21} encoders. We take the last-layer hidden states as representations of both the context images and prompts. The model is trained with a fixed learning rate of $1e$-$4$ for 50K iterations, using $256 \times 256$ images. We use a global batch size of $512$ for all runs.
Following prior works \cite{zhang2023adding, li2023gligen, wang2023context}, we apply random replacement of the prompts with empty strings for classifier-free guidance. We experimented with different rates and found that 50\% is the most optimal one, as also reported by ControlNet \cite{zhang2023adding}.
At inference time, we apply DDIM \cite{song2021denoising} as a default sampler, with 50 steps and a guidance weight of 3. Regarding the computational resources, the model is trained using 8 NVIDIA A100 GPUs.

\paragraph{Human Evaluation Setup.} To better quantify the performance, we perform a human evaluation to compare our model to Prompt Diffusion \cite{wang2023context}. A total of 10 in-house annotators participated in the study, annotating 240 randomly chosen samples from the test partition. For each test sample, we present two images - one generated by our Context Diffusion and another by Prompt Diffusion, randomly annotated as A and B, alongside the given visual context, query image, and prompt. Each annotator chooses either a preferred image or denotes both as equally preferred. We consider various in-domain and out-of-domain tasks for evaluation, across three distinct scenarios: using both visual context and prompts (C+P), using only visual context (C) and only prompt (P). Considering all these scenarios is highly important since it examines to what extent the models can learn from the conditional information in a balanced manner and whether they suffer when one input modality is not present.

\paragraph{Automated Evaluation.}
In addition to the human evaluation, we also use offline automated metrics to further evaluate the performance of Context Diffusion. In particular, we report FID \cite{heusel2017gans} scores for map-to-image tasks and RMSE scores for the image-to-map tasks, by taking the average across three random seeds. Note that we use 5000 randomly chosen test samples per setting for each task to generate output images both with our model and Prompt Diffusion.

\begin{figure}[t]
  \centering
    \includegraphics[width=\textwidth]{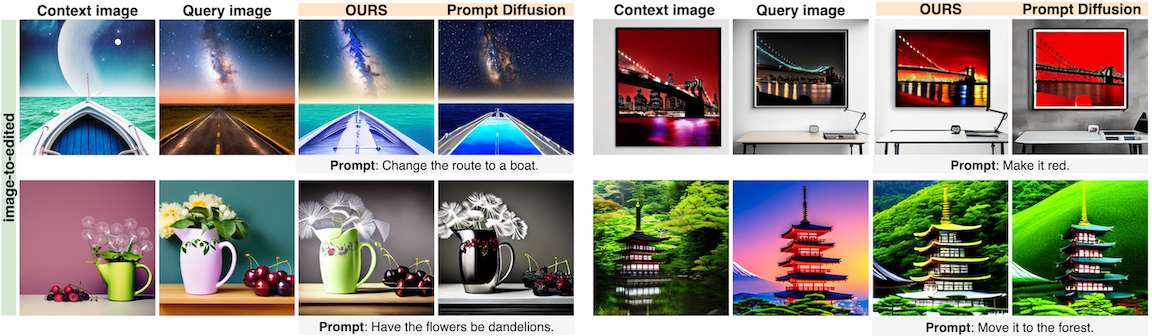}
    \vspace{-0.5cm}
  \caption{\textbf{Out-of-domain comparison to Prompt Diffusion~\cite{wang2023context}:} Image editing, with visual context and prompt (C+P) as conditioning information.}
  \label{fig:examples_edits}
  % \vspace{-0.5cm}
\end{figure} 
\begin{figure}[t]
   \centering
   \includegraphics[width=\textwidth]{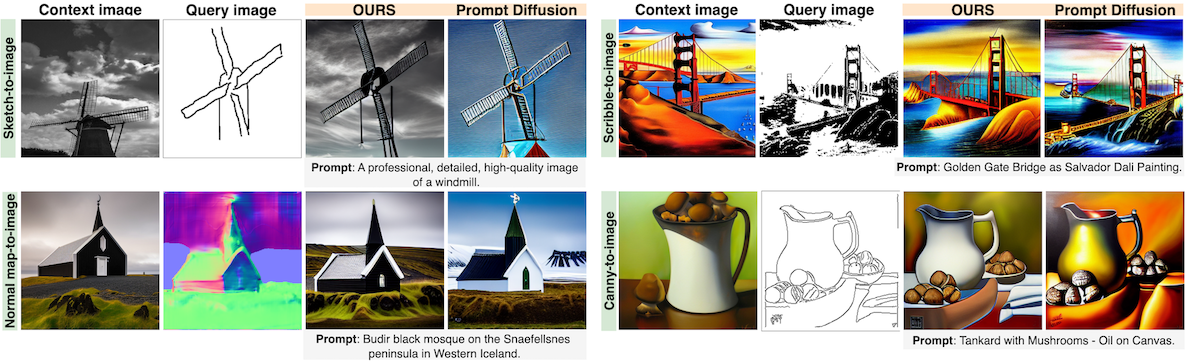}
   \vspace{-0.5cm}
   \caption{
        \textbf{Out-of-domain comparison to Prompt Diffusion~\cite{wang2023context}:} \{sketch, normal map, scribble, canny edge\}-to-image tasks.
        Visual context and prompt (C+P) are given as conditioning information.
    }
    \vspace{-0.5cm}
   \label{fig:out_domain_examples}
\end{figure}

\subsection{Results \& Discussion}
In this section, we compare our model against the most similar approach in the literature, \ie Prompt Diffusion~\cite{wang2023context}.
Prompt Diffusion expects a source-target pair of context images as an input, while in contrast our approach only requires context \ie target images. More analysis regarding this is provided in \ref{sec:ablations} Ablation study.
It is important to notice that in all comparisons we follow the source-target format of the input to generate images with Prompt Diffusion, but we omit the visualization of the source image from the figures to have consistent visualizations for both methods.
Additionally, both approaches operate with a query image and textual prompt as additional inputs.

We compare the methods across two important generalization axes: (i) \emph{in-domain} for seen and \emph{out-of-domain} for unseen tasks at training; (ii) visual context and prompt \emph{(C+P)}, context-only \emph{(C)}, and prompt-only \emph{(P)} variations of conditioning at inference time.
We also present the generated results of our model on \emph{few-shot} setup when several visual examples are given as input. Prompt Diffusion does not support the few-shot setup. Finally, we present the ablations for the key components of our proposed model.

\begin{table}[t]
    \setlength{\tabcolsep}{6pt}
    \centering
    \caption{
        \textbf{Human evaluation comparison to Prompt Diffusion (PD)~\cite{wang2023context}:} 
        In-domain and out-of-domain tasks, considering different conditioning settings: context image and prompt (C+P), visual context-image-only (C), prompt-only (P).
        We report the win rate as a percentage of winning votes for each model.
    }
    \vspace{-.2cm}
    \begin{tabular}{lcccccccc}
    \toprule
    & \multicolumn{4}{c}{\textbf{In-domain}} & \multicolumn{4}{c}{\textbf{Out-of-domain}} \\
    \cmidrule(lr){2-5} \cmidrule(lr){6-9}
    & C+P & C & P & \textit{avg} & C+P & C & P & \textit{avg}
    \\
    \midrule
        PD \cite{wang2023context} & 28.5 & 4.5 & \textbf{30.4} & 21.1 & 26.9 & 22.8 & 25.9 & 25.2 \\
        \rowcolor{Gray}
        \textbf{Ours} & \textbf{36.3} & \textbf{80.2} & 29.6 & \textbf{48.6} & \textbf{52.3} & \textbf{63.7} & \textbf{49.8} & \textbf{55.2} \\
    \bottomrule
    \end{tabular}
    \vspace{-0.5cm}
    \label{tab:human_eval_1_shot}
\end{table}

% \begin{wrapfigure}{L}{\textwidth}
%   \centering
%    \includegraphics[width=\textwidth]{figures/only_context_new.png}
%    \vspace{-0.5cm}
%    \caption{
%         \textbf{Conditioning comparison with Prompt Diffusion~\cite{wang2023context}:} Using visual context and prompt (C+P) and visual context-only (C) as conditioning, on both in-domain (image-to-HED) and out-of-domain (editing, scribble-to-image, sketch-to-image) tasks.
%     }
%    \label{fig:examples_no_prompts}
% \end{wrapfigure}

\begin{figure}[t]
  \centering
   \includegraphics[width=\textwidth]{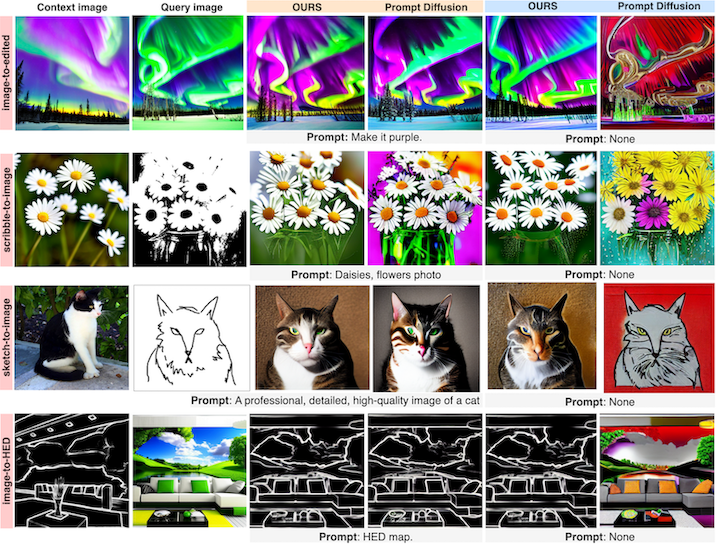}
   \vspace{-0.5cm}
   \caption{
        \textbf{Conditioning comparison with Prompt Diffusion~\cite{wang2023context}:} Using visual context and prompt (C+P) and visual context-only (C) as conditioning, on in-domain (image-to-HED) and out-of-domain (editing, scribble-to-image, sketch-to-image) tasks.
    }
   \label{fig:examples_no_prompts}
   \vspace{-0.5cm}
\end{figure}
% \vspace{-0.2cm}
\subsubsection{Data Domain}

\paragraph{In-domain Comparison.}
\vspace{-0.2cm}
We study the performance of models on the same data domain as the training data, but on the test data that is set aside.
This encompasses three ``forward'' tasks, \ie, the query image is either HED, segmentation, or a depth map while the expected output image is a real image, given the visual context and prompt in an adequate form.
Similarly, we evaluate three ``reverse'' tasks, where the query and output roles are reversed.
For the purpose of this discussion, we focus on the conditioning setup where both visual context and prompts \emph{(C+P)} are given as input.
Figure~\ref{fig:examples_in_domain} presents representative examples for each of the tasks: the first four columns depict the forward tasks, while the last four columns depict the reverse tasks.
It can be observed that our model can generate images with better fidelity to the context images and prompts, by managing to match the specific colors and styles from the context. On the other hand, Prompt Diffusion outputs are more saturated and fail to leverage the visual characteristics from the context (green radish instead of red in the second row). We include more examples in the supplementary materials.
These observations are further supported by the human evaluation presented in Table~\ref{tab:human_eval_1_shot} (In-domain (C+P) column), as well as in offline metrics comparison presented in Table~\ref{tab:fid_results} (C+P columns), where we obtain satisfactory performance improvement (36.3\% \vs 28.5\% win-rate) over Prompt Diffusion.

\paragraph{Out-of-domain generalization.}
% \vspace{-0.2cm}
The most advantageous aspect of having a model that is an in-context learner is its capacity to generalize to new tasks by observing the context examples given as input at inference.
Again, for the purpose of the discussion in this section, we focus on the conditioning setup where both visual context and prompt \emph{(C+P)} are given as input.
To test these generalization abilities, we consider tasks outside of the training domains: image editing with representative examples in Figure~\ref{fig:examples_edits}; \{sketch, normal map, scribbles, canny edge\}-to-image with representative examples in Figure~\ref{fig:out_domain_examples}.
In both figures, we observe noticeable improvements over Prompt Diffusion \cite{wang2023context}.
It is apparent that the visual characteristics of the context images are also transferred in the output images.
Furthermore, we select editing and sketch-to-image as representative out-of-domain tasks to perform a human evaluation study. We report the results in Table \ref{tab:human_eval_1_shot} (Out-of-domain (C+P) column), where we observe great improvements in win rate (52.3\% \vs 26.9\%), significantly higher than for in-domain setup, showing the advantage of in-context aware image generation.

\subsubsection{Conditioning variations at inference}

\paragraph{Using only visual context.}
\vspace{-0.2cm}
To better understand the effect of visual context examples on the model's performance, we analyze the outputs when the text prompt is not provided (empty string), \ie, only visual context is used as conditioning.
This experiment gives strong insights into the model's ability to perform in-context learning.
We show representative examples of this setup in Figure~\ref{fig:examples_no_prompts}.
It can be observed that Prompt Diffusion~\cite{wang2023context} is unable to learn from the visual examples, indicating that it relies solely on the text caption as conditional information.
We include this setting in the human evaluation and we report the results in Table~\ref{tab:human_eval_1_shot} ((C) columns). 
Overall, we observe a \emph{significant} performance gap between our model and Prompt Diffusion, both for in-domain (80.2\% \vs 4.5\% win-rate) and out-of-domain (63.7\% \vs 22.8\% win-rate) tasks. 
This result is additionally supported by the offline metrics in Table~\ref{tab:fid_results} ((C) columns) for in-domain tasks, further strengthening the observations that our model can truly leverage styles, color palettes, and object appearances in the visual context.

\begin{figure}[t]
  \centering
   \includegraphics[width=\textwidth]{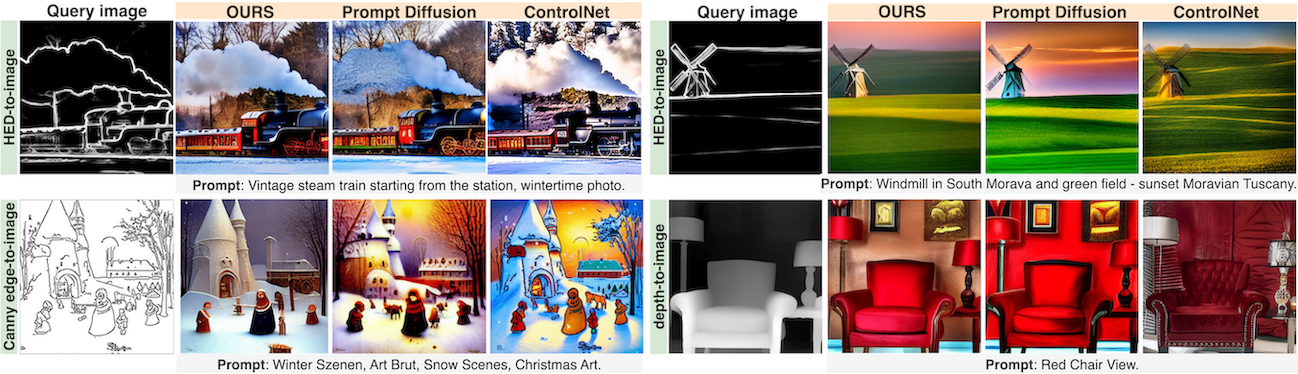}
   \vspace{-0.5cm}
   \caption{
        \textbf{Zero-shot comparison to ControlNet~\cite{zhang2023adding} and Prompt Diffusion~\cite{wang2023context}:} Using prompt-only (P) as conditioning information.
    }
   \label{fig:examples_no_context}
   % \vspace{-0.2cm}
\end{figure} 
\begin{table}[t]
    \setlength{\tabcolsep}{6pt}
    \centering
    \caption{
        \textbf{Offline comparison to Prompt Diffusion (PD)~\cite{wang2023context} using FID and RMSE:}
        In-domain tasks across three different conditioning settings: visual context and prompt (C+P), visual context-only (C), prompt-only (P).
        Lower scores are better.
    }
    \vspace{-0.3cm}
    \begin{tabular}{lccccccccc}
    \toprule
    & \multicolumn{9}{c}{\textbf{FID (map-to-img)} $\downarrow$ } \\
    & \multicolumn{3}{c}{\textbf{HED-to-img}} & \multicolumn{3}{c}{\textbf{seg-to-img}} & \multicolumn{3}{c}{\textbf{depth-to-img}} \\
    \cmidrule(lr){2-4} \cmidrule(lr){5-7} \cmidrule(lr){8-10} 
    & \small{C+P} & \small{C} & \small{P} & \small{C+P} & \small{C} & \small{P} & \small{C+P} & \small{C} & \small{P} 
    \\
    \midrule
        PD \cite{wang2023context} & 12.8 & 22.5 & 15.1 & 16.7 & 25.1 & \textbf{17.2} & 15.9 & 27.0 & 18.1 \\
        \rowcolor{Gray}
        \textbf{Ours} & \textbf{12.3} & \textbf{17.7} & \textbf{14.8} & \textbf{13.4} & \textbf{19.0} & 18.5 & \textbf{12.9} & \textbf{18.5} & \textbf{17.5} \\
    \midrule
    & \multicolumn{9}{c}{\textbf{RMSE (img-to-map)} $\downarrow$ } \\
    & \multicolumn{3}{c}{\textbf{img-to-HED}} & \multicolumn{3}{c}{\textbf{img-to-seg}} & \multicolumn{3}{c}{\textbf{img-to-depth}} \\
    \cmidrule(lr){2-4} \cmidrule(lr){5-7} \cmidrule(lr){8-10} 
    & \small{C+P} & \small{C} & \small{P} & \small{C+P} & \small{C} & \small{P} & \small{C+P} & \small{C} & \small{P}
    \\
    \midrule
        PD \cite{wang2023context} & 0.15 & 0.33 & \textbf{0.15} & 0.32 & 0.41 & 0.32 & \textbf{0.14} & 0.34 & 0.14 \\
        \rowcolor{Gray}
        \textbf{Ours} & \textbf{0.11} & \textbf{0.11} & 0.16 & \textbf{0.29} & \textbf{0.28} & \textbf{0.30} & \textbf{0.14} & \textbf{0.13} & \textbf{0.13} \\
    \bottomrule
    \end{tabular}
    \vspace{-0.4cm}
    \label{tab:fid_results}
\end{table}

\paragraph{Using only text prompts.}
\vspace{-0.2cm}
Apart from being able to handle scenarios only with visual context, we aim to also support scenarios using only text prompts. 
To enable this setting, we simply mask out the visual context by using black images.
This essentially yields zero-shot settings, boiling down to how ControlNet~\cite{zhang2023adding} is used at inference time. 
However, unlike ControlNet which requires a separate model trained for each task, our Context Diffusion generalizes across a series of tasks.
In Figure \ref{fig:examples_no_context} we show representative examples of this setting, comparing our model to ControlNet and Prompt Diffusion.
It can be seen that our model can generate more realistic images compared to ControlNet and Prompt Diffusion. 
Similar to before, we also include this setting in the user study, reported in Table~\ref{tab:human_eval_1_shot} ((P) columns). We observe much better performance on the out-of-domain (49.8\% \vs 25.9\%) tasks compared to Prompt Diffusion, and a slight decrease on in-domain (29.6\% \vs 30.4\%) tasks.
% We observe a less than 1 point decrease in performance compared to Prompt Diffusion on in-domain (29.6\% \vs 30.4\%) and much better on out-of-domain (49.8\% \vs 25.9\%) tasks.
This supports our observations that Prompt Diffusion relies too much on the textual prompt, as well as suffers in out-of-domain data regimes.
Further, we compare the automated metrics in Table~\ref{tab:fid_results} ((P) columns), again observing better overall performance across different tasks.

\begin{figure}[t]
   \centering 
   \includegraphics[width=\textwidth]{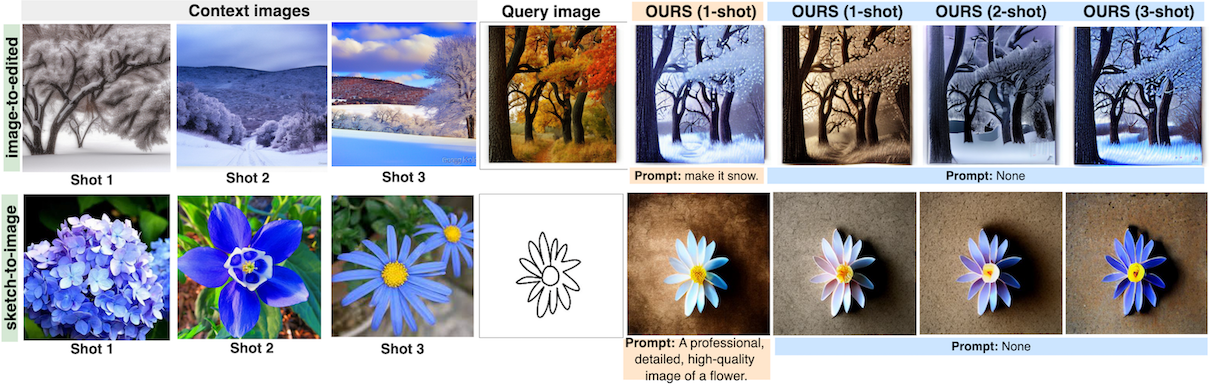}
   \vspace{-0.5cm}
   \caption{\textbf{Few-shot examples:} Comparison between out-of-domain tasks (editing and sketch) using one context example with a text prompt, and one, two, and three shots of context examples with no text prompt. Our model can leverage multiple visual examples to handle scenarios when the text prompt is not present. 
   }
   \label{fig:examples_k_shots}
   \vspace{-0.5cm}
\end{figure}
\begin{wraptable}{r}{5.5cm}
\vspace{-1.2cm}
\caption{
        \textbf{Human evaluation for 1-shot \vs 3-shot setups:} 
        Out-of-domain tasks, considering different conditioning settings: visual context and prompt (C+P), visual context-only (C).
        Note that the ``prompt-only'' (P) setting corresponds to a zero-shot scenario and is not applicable here.
        We report the win rate as a percentage of winning votes for each model.
    }
    \setlength{\tabcolsep}{3pt}
    \begin{tabular}{lccc}
    \toprule
    & \multicolumn{3}{c}{\textbf{Out-of-domain}} \\
    \cmidrule(lr){2-4}
    & C+P & C & \textit{avg}
    \\
    \midrule
        Ours (1-shot) & 21.5 & 28.3 & 24.9 \\
        \rowcolor{Gray}
        \textbf{Ours (3-shot)} & \textbf{60.0} & \textbf{50.2} & \textbf{55.1} \\ 
    \bottomrule
    \end{tabular}
    \vspace{-0.6cm}
    \label{tab:human_eval_13_shot}
\end{wraptable}  

\subsubsection{Few-shot visual context examples}
Context Diffusion is designed to handle multiple context examples, enabling few-shot scenarios.
Using one context example proved to be enough for in-domain tasks, as seen in Figure~\ref{fig:examples_in_domain}, even without using a text prompt as seen in Figure \ref{fig:examples_no_prompts}.
Therefore in the few-shot experiments, we focus on the out-of-domain tasks, such as editing and sketch-to-image.
In particular, we augment the visual context with additional images, depicting similar objects, scenes, or desired color palettes.
Moreover, we look at scenarios where the textual information is not present since in that case, the model has to rely on the visual context only.
As can be seen from Figure~\ref{fig:examples_k_shots}, multiple demonstrations of images help to strengthen the target visual representation, especially when the prompt is not present.
We also quantify the performance by conducting a human evaluation for the few-shot settings, presented in Table \ref{tab:human_eval_13_shot}.
We compare our model when using one context example \vs using three examples.
Overall we observe improved performance (55.1\% \vs 24.0\% average win rate) when using three context images which aligns with the qualitative observations. 
In the current experiments, we use 1 up to 3 shots as a representative few-shot setting, however, our model can accommodate more than 3 shots.

\subsubsection{Ablations}
\label{sec:ablations}
\begin{figure*}[t!]
    \centering
    \begin{subfigure}[t]{0.48\textwidth}
        \centering
        \includegraphics[width=\textwidth]{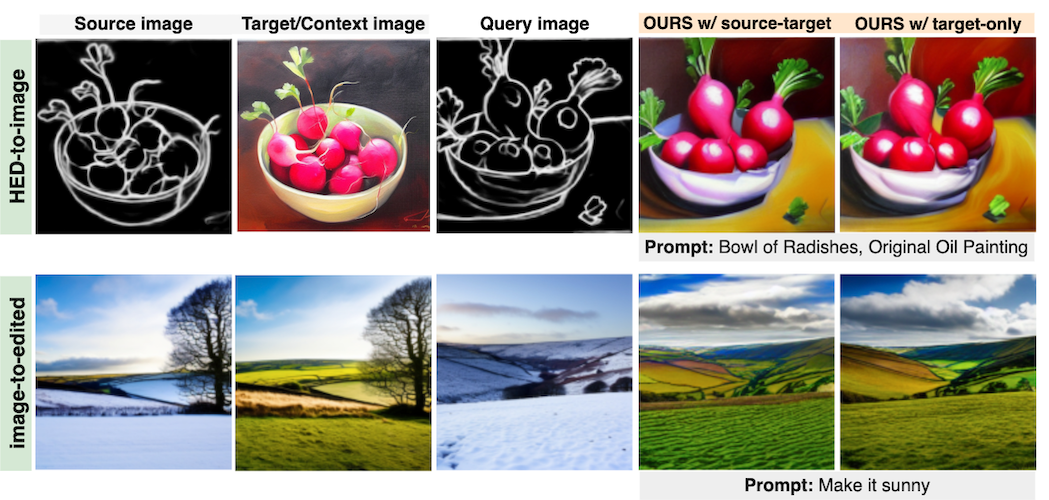}
       \caption{\textbf{Source-target vs target-only as context.} Having a source image paired with the target does not influence the output image, since the visual information is provided by the target only.}
       \label{fig:ablation_source_target}
    \end{subfigure}%
    ~ 
    \begin{subfigure}[t]{0.48\textwidth}
        \centering
        \includegraphics[width=0.85\textwidth]{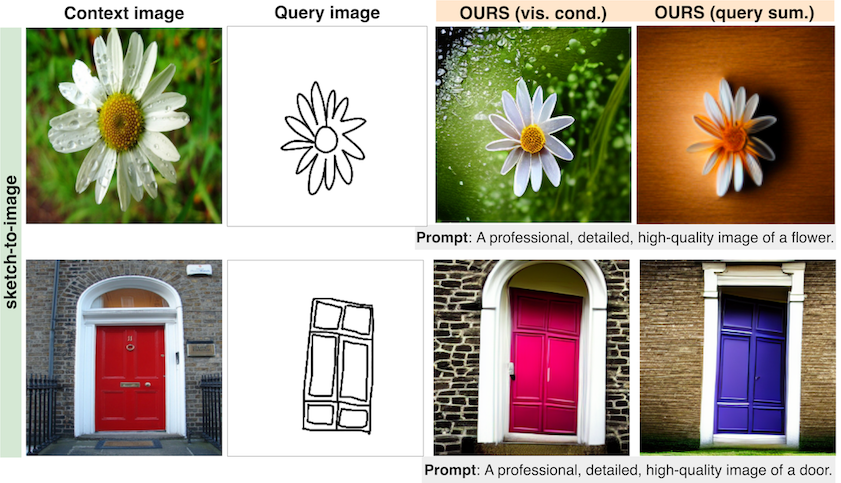}
   \caption{\textbf{Benefit of visual conditioning.} 
   Ingesting the context examples as visual conditioning helps in leveraging the visual cues, unlike the summation of the context to the query image. }
   \label{fig:ablation_1_modelling}
    \end{subfigure}
    \caption{\textbf{Ablations.} We ablate the key components of our framework, namely (a) Using source-target vs target-only as context, (b) Benefit of visual conditioning. }
    \vspace{-0.5cm}
\end{figure*}
\vspace{-0.3cm}

\paragraph{Source-target vs target-only as context.}
We analyze the performance when training by using source-target image pairs as context examples (same as Prompt Diffusion \cite{wang2023context}). Figure \ref{fig:ablation_source_target} shows that there is no benefit when the source image is paired with the target image during the training. The visual information needed for generating the image is entirely contained in the target \ie context image or the prompt, while the query image controls the layout. 

\paragraph{Benefit of visual conditioning.}
Our model employs a visual conditioning pipeline to ingest the context examples similarly to the text prompts. To better understand the benefit of such conditioning, we compare it to the same paradigm used in Prompt Diffusion \cite{wang2023context}. Specifically, we use ConvNets to encode the context examples and then we sum them to the query image. The entire representation is then fed into the ControlNet module. As we can see in Figure \ref{fig:ablation_1_modelling}, training in this manner prevents the model from actually leveraging the context.

\section{Conclusion}
\label{sec:conclusion}
We present an in-context-aware image generation framework capable of learning from a variable number of visual context examples and prompts. 
Our approach leverages both the visual and text inputs as conditioning information, resulting in a framework able to learn in a balanced manner from the multimodal inputs. Furthermore, learning from a few context examples showed to be helpful in learning strong visual characteristics, especially if the prompt is not available. 
Our experiments and human evaluation study demonstrate the applicability of our approach across diverse tasks and settings, confirming the improved quality and context fidelity over prior work.


\bibliographystyle{splncs04}
\bibliography{main}

\end{document}